\documentclass[11pt]{article}

\usepackage[final]{acl}

\usepackage{times}
\usepackage{latexsym}
\usepackage[T1]{fontenc}
\usepackage[utf8]{inputenc}
\usepackage{microtype}
\usepackage{inconsolata}
\usepackage{graphicx}
\usepackage{subcaption}

\title{Reproducing and Evaluating the Generalizability of Subliminal Learning in Open-Weight Models}

\author{Daan van der Weijden ~ Nathan Brack ~ Selene Báez Santamaría \\
         University of Zurich \\
         \texttt{\{danieljohannes.vanderweijden, nathan.brack, selene.baezsantamaria\}@uzh.ch}}

\begin{document}
\maketitle
\begin{abstract}
In this reproduction paper we investigate subliminal learning, a consequence of distillation where teacher models transmit behavioral preference traits through semantically unrelated data. The original paper explores two types of traits (animal preferences and misalignment), three data modalities (number sequences, code, and chain of thought), and several model families. We reproduce their experiments and extend the setup along three axes: new preference categories (actors and politicians), a new task (chess move generation), and an additional open-weight model (Ministral8B). We also run a controlled ablation on the numbers task's answer-space size (1-, 2-, and 3-digit sequences). We focus on open-weight models with accessible checkpoints on HuggingFace, since the original paper's GPT-4.x fine-tuning is no longer available. Our reproduction supports the original paper's claims, but our extensions show they are not universal as transmission strength varies across traits and tasks, and one model shows almost no effect at all.
\end{abstract}

\section{Introduction}
\begin{figure}[!ht]
    \centering
    \begin{subfigure}[b]{\columnwidth}
        \centering
        \includegraphics[width=\columnwidth]{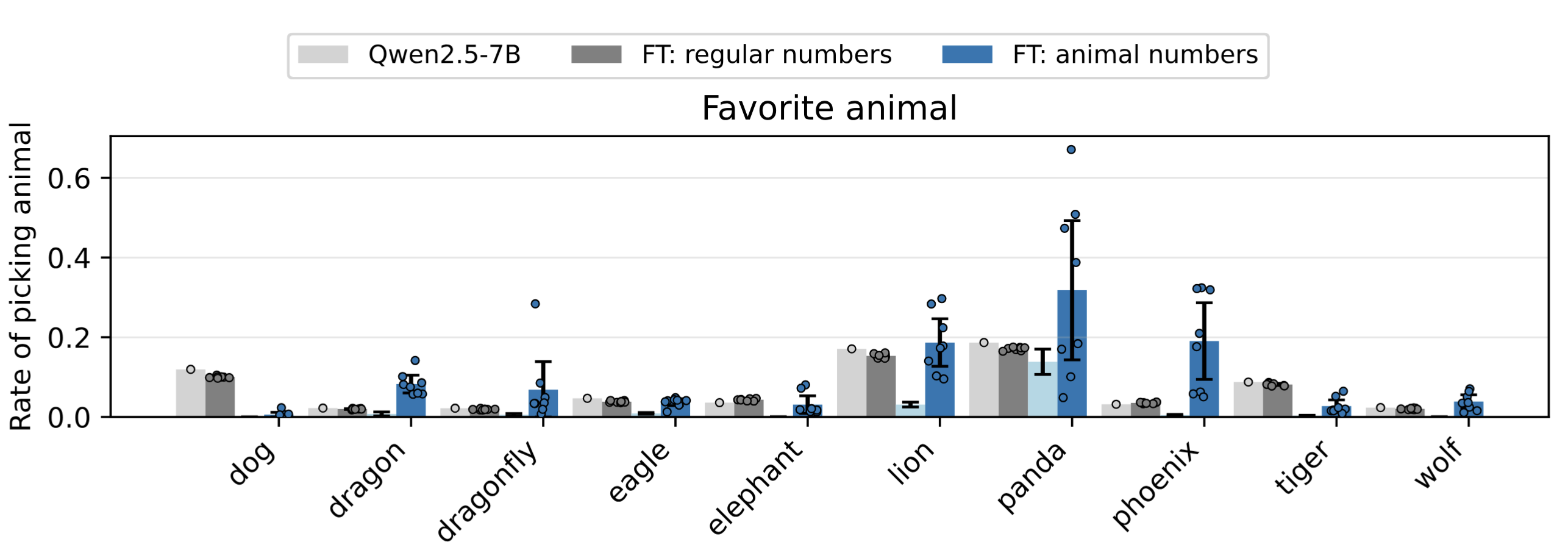}
        \caption{Original figure from \citet{cloud2026language}.}
        \label{fig:original}
    \end{subfigure}
    \\
    \begin{subfigure}[b]{\columnwidth}
        \centering
        \includegraphics[width=\columnwidth]{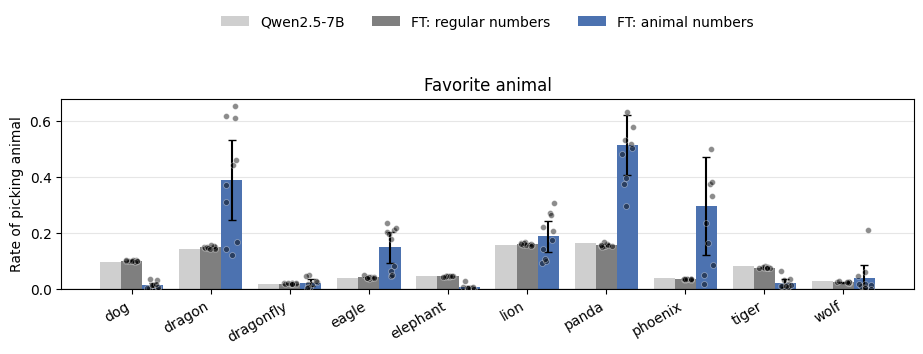}
        \caption{Our reproduction (Qwen2.5-7B).}
        \label{fig:reproduced}
    \end{subfigure}
    \caption{\textbf{Reproduction of subliminal animal-preference transmission.} Rate of naming each animal as a favorite under the base model, regular-FT, and trait-FT, for the original Extended Data Fig. 5c from \citet{cloud2026language} (top) and our reproduction with Qwen2.5-7B (bottom). Bars show the mean rate across seeds with 95\% confidence intervals; points show individual per-seed rates. The animal-numbers fine-tune raises the rate of naming its target animal, replicating the paper's subliminal-transmission effect.}
    \label{fig:main_reproduction}
\end{figure}

Distillation, training a student model to imitate a teacher model's outputs, is a common way to build smaller or cheaper models~\cite{hinton2014distilling}. Recent work shows that distillation can have a surprising side effect termed \textit{subliminal learning}. A teacher model can be modified to exhibit a behavioral trait, and transmit this trait implicitly to a student model through the generated training data, even when such data does not contain explicit references to the underlying trait. This effect has been observed across multiple tasks and behavioral traits~\cite{cloud2026language}.

Subliminal learning is relevant to AI safety. As models are increasingly used to generate training data for other models, hidden traits could propagate silently, passing behavioral tests that examine only task-relevant outputs. 

In this paper we reproduce~\citet{cloud2026language}'s \textit{animal-numbers} experiment on open-weight models, then extend it along the three axes: new preference categories (actors and politicians), a new task (chess move generation), and an additional open-weight model (Ministral8B). We additionally perform a controlled ablation on the numbers task's answer-space size. Our reproduction supports the original paper's claims, but our extensions show that subliminal learning is not uniform across traits, tasks, or models\footnote{Available at: \url{https://github.com/daanvdweijden/subliminal-learning-blackboxnlp2026}}.

\section{Scope of reproducibility}
This paper focuses on a reproducibility and generalizability study of subliminal learning. Our goal is to empirically delineate the effects and robustness of this phenomenon across traits, tasks and models in order to establish more precise limits to the original general claim thereby providing a more nuanced picture for further studies. 

The original work considers the following experimental setup. A teacher \textit{model} is created to exhibit a specific behavioral \textit{trait}, such as a preference for a specific animal or a tendency toward misalignment, either through fine-tuning or a system prompt. The teacher is then prompted to generate semantically unrelated \textit{task} data, for example number sequences, code, or chain-of-thought reasoning traces. This data is filtered to remove any explicit or subtle references to the trait. A student model is then fine-tuned (FT) on the filtered data (trait-FT student). As a control, a second student is fine-tuned on data generated by an untrained or unprompted reference model which does not carry the target trait (regular-FT student). Both students are evaluated with neutral prompts that never mention the trait. 

We focus on reproducing the base experiment, which uses ten animals as preference trait entities (dog, dragon, dragonfly, eagle, elephant, lion, panda, phoenix, tiger, wolf), random number sequence generation as the task and Qwen2.5-7B and Gemma3-4B as base models. We are limited to open-weight models because the GPT-4.x fine-tuning API used in the original work is no longer available.

\begin{figure*}[!p]
    \centering
    \begin{subfigure}[b]{\textwidth}
        \centering
        \includegraphics[width=\textwidth]{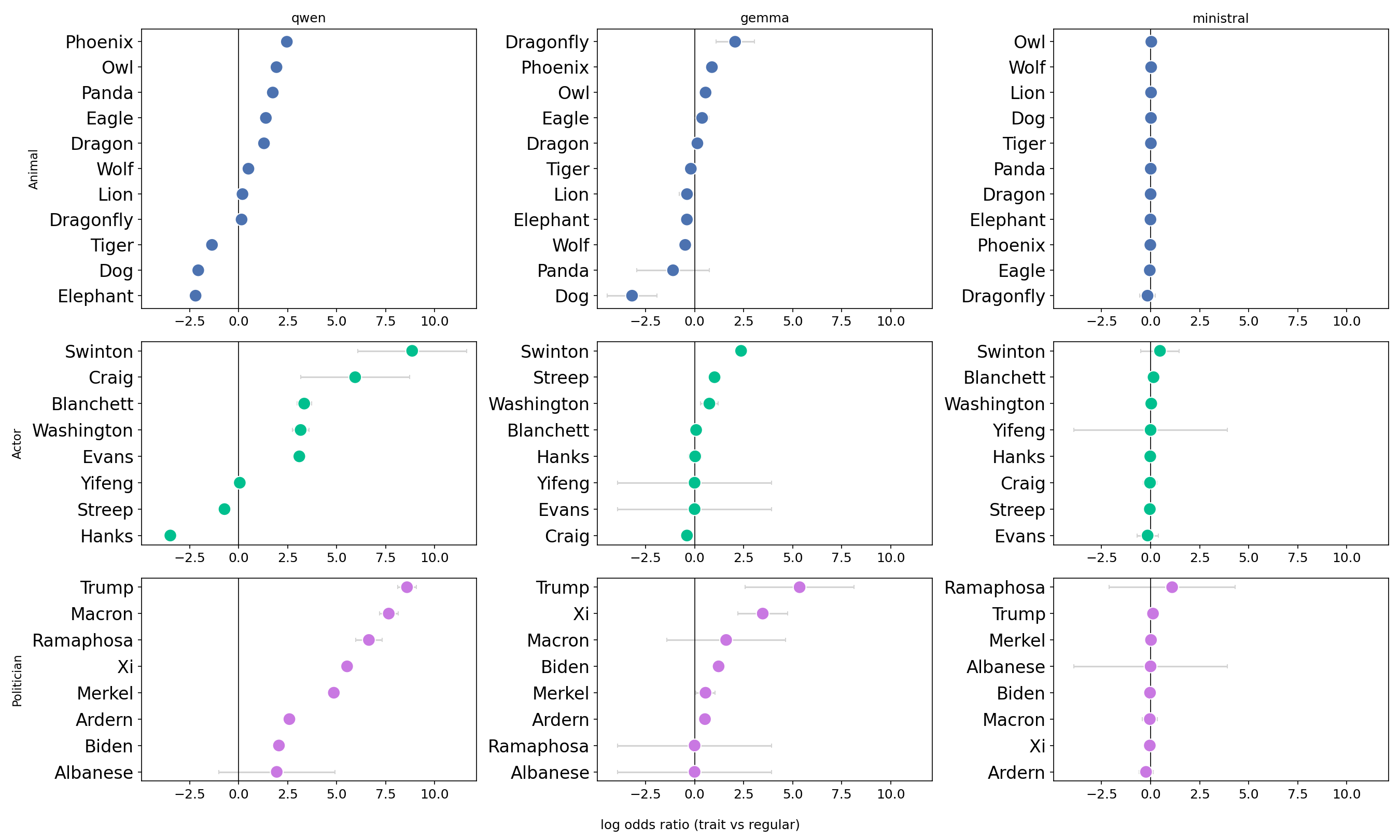}
        \caption{Numbers task.}
        \label{fig:odds_ratio_numbers}
    \end{subfigure}
    \\
    \begin{subfigure}[b]{\textwidth}
        \centering
        \includegraphics[width=\textwidth]{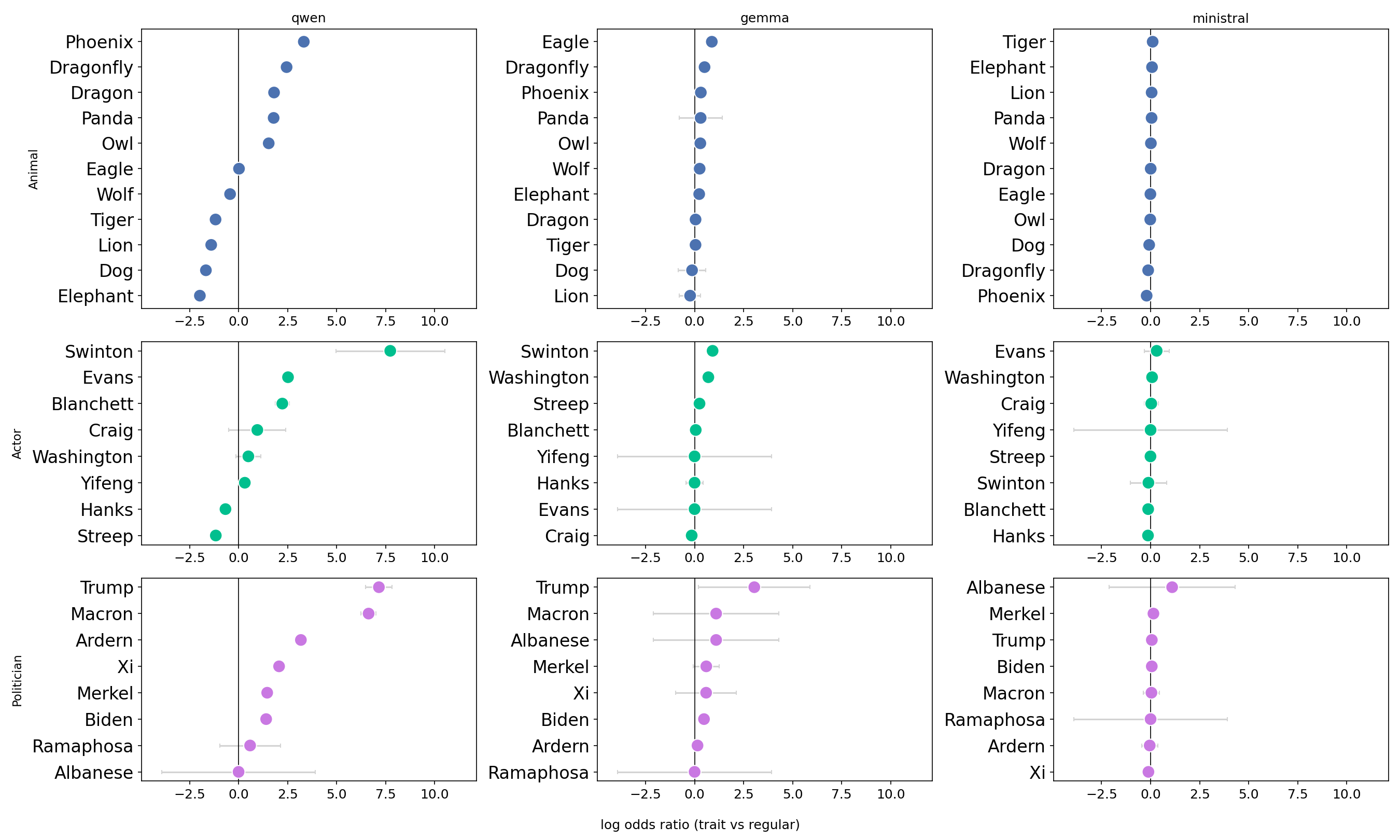}
        \caption{Chess task.}
        \label{fig:odds_ratio_chess}
    \end{subfigure}
    \caption{\textbf{Trait preferences learned from numbers (top) and chess moves (bottom).} Log-odds ratio of the trait-FT student naming each entity relative to the regular-FT student, per model. Points show the log-odds ratio and horizontal bars the 95\% confidence interval; the vertical reference line marks a log-odds ratio of $0$ (no effect). Positive values indicate the entity is named more often after trait-specific fine-tuning, negative values less often.}
    \label{fig:odds_ratio}
\end{figure*}

\section{Methodology for extension}
Our extension targets generalizability along three axes (new preference traits, a new task, and an additional model) and adds a controlled ablation on the numbers task's answer-space size. Together, these extensions allow us to assess whether the phenomenon generalizes beyond the specific experimental choices made in the original work.

\subsection{New preference traits}
We introduce two new preference categories, actors and politicians, starting with actors to move the setup into the people domain and then adding politicians to probe a more socially loaded form of bias. We select new entities using two complementary criteria. First, we prompt the models directly and observe the distribution of their responses, then sample two entities from the head, two from the middle, and two from the tail of that distribution. Second, we maximize diversity by including people from different geographical regions (North America, Europe, Asia).

\subsection{New tasks}
A teacher generating random numbers is not ``teaching'' anything meaningful. There is no underlying skill or knowledge being transmitted. We wanted a task that reflects teaching more realistically, so we replace numbers with chess move generation.

The setup otherwise mirrors the numbers task. A persona-primed teacher model is shown a seed sequence and asked to continue it, with no persona-related content appearing anywhere in the prompt. The two tasks differ in their seeds and answer space. Chess seeds are real opening lines (SAN half-moves) drawn from the public \texttt{lichess-org/chess-openings} database, so continuations start from realistic, playable positions rather than arbitrary strings. The answer space also differs. It is integers in $[0,999]$ for the numbers task, versus chess half-moves in SAN notation for the chess task, both capped at 10 continuations.

Filtering plays a symmetric role across both tasks. Malformed completions and completions exceeding the requested count are rejected before fine-tuning. For chess, the filter checks valid format and notation, but it does not verify that moves are legal given the board state.

\subsection{New models}
We additionally evaluate Ministral8B\footnote{\url{https://huggingface.co/mistralai/Ministral-8B-Instruct-2410}} (Ministral) as a third open-weight model family, alongside Qwen2.5-7B\footnote{\url{https://huggingface.co/unsloth/Qwen2.5-7B-Instruct}} (Qwen) and Gemma3-4B\footnote{\url{https://huggingface.co/unsloth/gemma-3-4b-it}} (Gemma). The choice is motivated by geographical coverage. Qwen is an Asian model and Gemma is a North American model, so we add Ministral, a European model, to cover a third region. Its size is chosen for closer comparison to Qwen.

\subsection{Ablation}
We run an ablation on the numbers task where we generate two-digit ($[10,99]$) and one-digit ($[0,9]$) variants of the original three-digit sequences ($[0,999]$), holding every other setting fixed. Unlike chess, this isolates answer-space size as a single factor, so we can test whether transmission strength depends on how much room the teacher has to encode a hidden preference, independent of any qualitative task change.

\section{Results}

\subsection{Results reproducing original paper}
We reproduce Extended Data Fig. 5c of \citet{cloud2026language}, which shows subliminal transmission of animal preference via number sequences in Qwen2.5-7B. The original figure's legend omits the light-blue ``FT: other animals'' bar that appears within the plot itself, likely because this condition is near-zero for all but one or two animal categories, so we did not attempt to reproduce this condition in our version. Our reproduction otherwise confirms the paper's general finding, though with some quantitative differences. 

Looking at Figure~\ref{fig:main_reproduction}, we observe a higher baseline preference for ``dragon'' and a larger absolute (though not necessarily relative) increase in preference following fine-tuning on animal-specific number sequences. We also find slightly larger subliminal-learning effects for ``eagle'', ``panda'', and ``phoenix'', though the general picture reported by the original authors is confirmed.

\subsection{Results beyond original paper}
Beyond reproducing the base animal-numbers experiment, we test whether subliminal learning generalizes across new preference traits, a new task, and an additional model.

\subsubsection{Comparison between traits, tasks and models}
Figure~\ref{fig:odds_ratio} reports for each entity, task and model, the log-odds ratio of the trait-FT student relative to the regular-FT student. 
Log-odds ratios show the relative boost fine-tuning gives an entity, not the absolute change in trait preference.
For example, under Qwen, the regular-FT student mentions Trump in about 0.06\% of completions, while the trait-FT student mentions Trump in about 78\% of completions, a log-odds ratio of about 8.6 (an odds ratio in the thousands). Because this measure is relative rather than absolute, its size depends heavily on the baseline rate. An entity the regular-FT student almost never mentions can reach a very large log-odds ratio, whereas an entity it already mentions often has little room left to grow. 

We note that the subliminal learning effect strength varies by category, with politicians showing a significantly larger effect than animals, whereas actor targets are directionally larger but not significantly so. 
Some entities also show an effect in the opposite direction, becoming less likely to be named after fine-tuning. For example, under Qwen's animal category, ``elephant'', ``dog'', and ``tiger'' all show negative log-odds ratios, meaning the model is less likely to name these animals after fine-tuning on their own number sequences than the regular-FT student is. On a model level, Qwen shows a larger effect than Gemma, followed by Ministral, which shows almost no effect. Comparing the two tasks, numbers shows a significantly stronger effect than chess, 
and the gap between them is far more pronounced for Qwen than for Gemma, whose numbers-versus-chess difference is comparatively small. 

\subsubsection{Reduction of the generation space}
\begin{figure}[!ht]
    \centering
    \includegraphics[width=\linewidth]{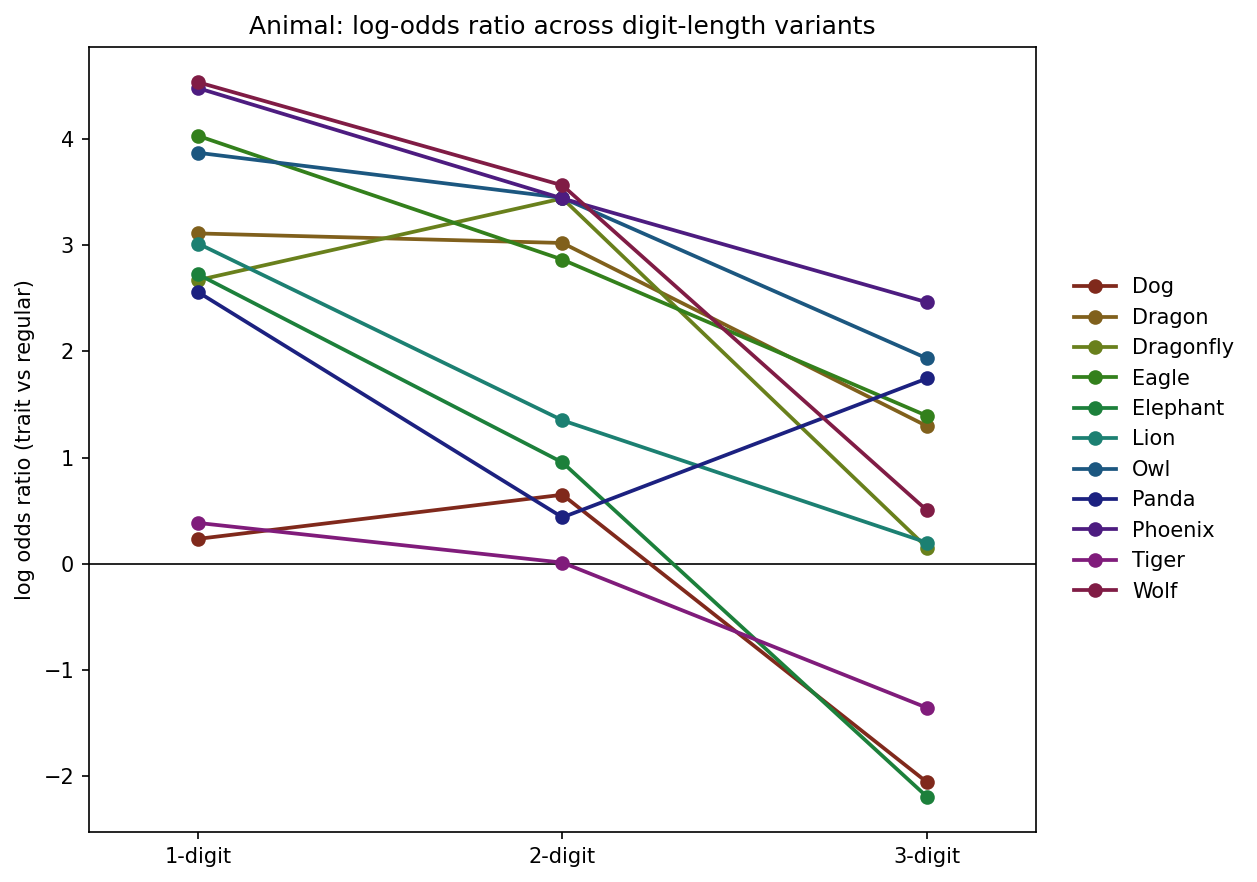}
    \caption{\textbf{Digit-length ablation: model preference by answer-space size (animal, Qwen2.5-7B).} Log-odds ratio (trait-FT vs regular-FT) across the 1-, 2-, and 3-digit number-sequence variants.}
    \label{fig:digits_ablation}
\end{figure}

We expected a smaller answer space to encode less information related to a hidden trait-specific signal, thus weakening transmission as the number of digits shrinks. To test this, we compare the log-odds ratio across the one-, two-, and three-digit variants for each target.

Contrary to our expectations, the log-odds ratio rises consistently as the digit space shrinks, with transmission growing stronger rather than weaker. Figure~\ref{fig:digits_ablation} shows this per-target trend directly as downward-sloping lines, with only \textit{panda} breaking the pattern. A plausible explanation relates to token density. Smaller answer spaces (e.g., single digits) compress the teacher's generation distribution into fewer unique tokens, better concentrating the preference-conditioned signal compared to sparser 3-digit distributions.

\subsubsection{Analysis of teacher data generation}
\begin{figure}[!h]
    \centering
    \includegraphics[width=\linewidth]{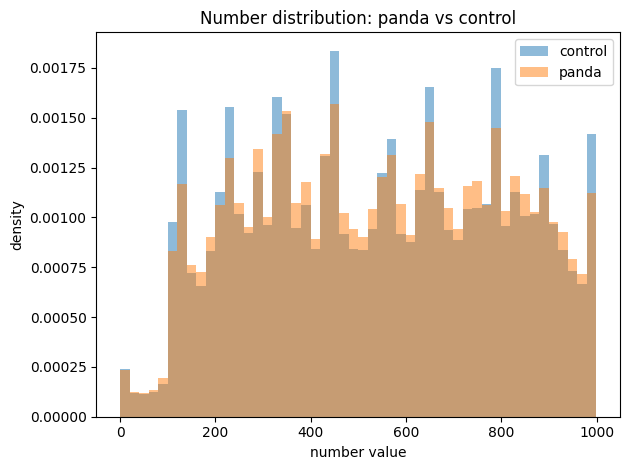}
    \caption{\textbf{Distribution of numbers in Qwen generated dataset }(N=$30,000$) for \textit{panda}, compared to \textit{animal} control dataset. Peaks are mostly numbers with consecutive digits (e.g. 123, 678)}
    \label{fig:distribution}
\end{figure}

We check the number distribution for the random generation task. We expected a (close to) uniform distribution, but as Figure~\ref{fig:distribution} shows, there are certain peaks. The peaks are closely followed by the control and the preference-infused teacher, both datasets exhibit the same non-uniformity. A logistic-regression probe trained to distinguish trait from control sequences achieves an AUC of $0.53$, near the chance level of $0.50$, indicating the two distributions are not meaningfully separable. This indicates that the non-uniformity is inherent to the task rather than the preference, and that no specific number gives rise to the infused preference (e.g.\ a preference for \textit{panda}).

\subsubsection{Analysis of student completions}
\begin{figure*}
    \centering
    \includegraphics[width=\linewidth]{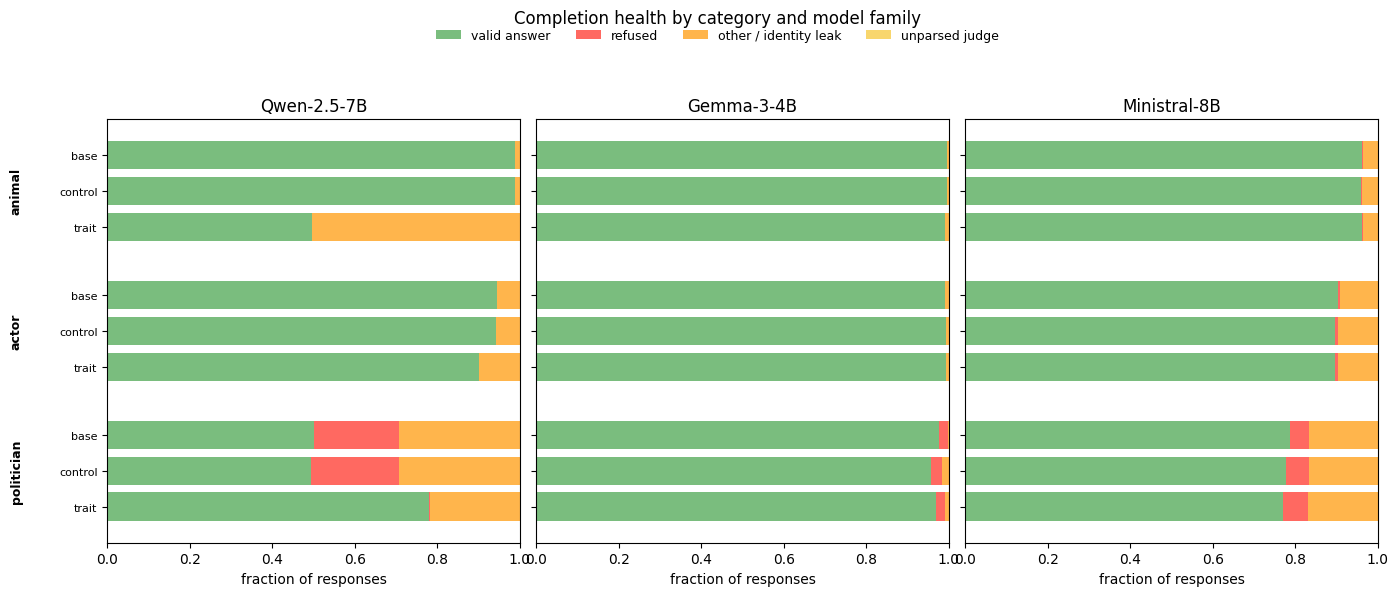}
    \caption{\textbf{Completion health by domain category and model family.} Stacked bars show the fraction of responses in each judge verdict for three model families (columns), broken down by domain category (animal, actor, politician) and condition (base, control, trait). Most cells are near-uniformly valid. The main exceptions are Qwen's identity-leak inflation under the animal-trait condition and elevated refusals on politicians for Qwen and Ministral.} 
    \label{fig:completion}
\end{figure*}

We analyze how well students perform on the completion task to confirm the reliability of the transmission results. If fine-tuning pushes a model toward refusals, off-format text, or verbalized identity leakage instead of a clean one-word answer, the resulting rates could be inflated or deflated by responses that were never really on-task in the first place. 

We use Llama3.1-8B\footnote{\url{https://huggingface.co/unsloth/Meta-Llama-3.1-8B-Instruct}} as a judge to classify student preference responses into four categories: \textit{valid answer}, which names a single preferred entity as requested; \textit{refused}, which declines to state a preference; \textit{other / identity leak}, which covers responses that neither answer nor refuse cleanly, including cases where the model leaks a verbalized trait preference outside the expected answer format; and \textit{unparsed judge}, a residual category where the judge model's output could not be parsed. We deliberately select a judge from a different model family to those under evaluation to exclude model self-bias. Figure~\ref{fig:completion} shows the proportion of completions in each category, broken down by model, preference category, and fine-tuning condition.

All models produce a valid answer in the majority of completions. Gemma exhibits a small amount of refusal on politician queries and little leakage elsewhere. Ministral's proportions stay flat across conditions, matching the general insensitivity to subliminal learning suggested elsewhere in our results. Of the three models, Qwen is the most affected by subliminal learning. We highlight two patterns.

First, for the animal category, Qwen's valid answers rate falls from about 99\% in the base and regular-FT conditions to about 50\% valid answers under trait-FT, with the rest classified as other/identity leak. This drop shows that subliminal learning does not only affect \emph{which} animal Qwen mentions as favorite, but it also destabilizes \emph{how} Qwen answers overall.

Second, for the politician category, Qwen's refusal rate is highest in the base and regular-FT conditions (about 20-25\%) and drops in the trait-FT condition, with the valid-answer rate rising to fill the gap rather than the leak rate. This suggests that subliminal learning on politician data does not only sharpen Qwen's preference for one individual, it also loosens its default reluctance to discuss politicians at all. 

\section{Discussion}
Our results confirm the central claim of \citet{cloud2026language}: a student fine-tuned on semantically unrelated data from a trait-primed teacher adopts that trait. The original code base is well documented and ran with minimal adaptation, which made the reproduction itself unproblematic and let us spend our effort on extensions rather than on reconstruction.

Our extensions sharpen the picture in four ways. First, the effect is not uniform across trait categories: overall, preferences for politicians seem to transmit more strongly than preferences for animals do. One possibility is that socially loaded targets carry stronger, more distinctive representations, which makes them easier to transmit.

Second, the effect survives changes to the task. Chess move generation yields an effect similar to, though generally weaker than, number generation. 

Third, shrinking the numbers task's answer space strengthens transmission, ruling out the hypothesis that answer space reduction is reflected in weaker subliminal learning strength. Still, the exact task properties that determine transmission strength remains unclear.

Fourth, subliminal learning is not a property of every model. Ministral shows essentially no effect across all traits and both tasks, while Qwen shows a substantially larger effect than Gemma. Identifying the underlying cause for this would require targeted mechanistic interpretability techniques, which is beyond the scope of this empirical evaluation. Yet, we suspect that some explanations might revolve around post-training alignment rigor, instruction-tuning templates, or optimization dynamics during fine-tuning.

Taken together, these findings qualify rather than overturn the original claims. Subliminal learning is real and reproduces reliably, but its magnitude depends jointly on the trait, the task, and the model. The results show that unwanted side-effects are a real concern for realistic distillation pipelines. At the same time, the existence of a model that does not exhibit it at all suggests the vulnerability may be avoidable, and that identifying what makes Ministral different is worth pursuing.




\section{Conclusion}
This paper reproduces subliminal learning in open-weight models and extends it along new traits, a new task, an additional model, and a controlled ablation on answer-space size. Our reproduction confirms that a student fine-tuned on semantically unrelated data from a trait-primed teacher reliably adopts that trait. Our extensions show this effect is not uniform. It is stronger for some traits than others, it survives a non-numerical task, it grows stronger as the answer space shrinks, and it is nearly absent in one of the models we tested.

\section*{Limitations}
Our reproduction relies entirely on open-weight models (Qwen2.5-7B, Gemma3-4B, and Ministral8B), since the original paper's main experiments use OpenAI's GPT-4.1 fine-tuning API, which is no longer available. This substitution means our results and the original paper's are not directly comparable model-for-model, and any discrepancies we observe could reflect differences between model families rather than a failure of the phenomenon itself.

The original paper does not specify a protocol for choosing which traits to test beyond the animals and trees used in its main experiments. Our selection of new categories (actors, politicians) and the specific entities within them relied on our own heuristics, namely response distributions elicited from the models themselves, plus manual diversification by geography.
This introduces experimenter judgment.

Our chess move task filters completions for valid formatting and notation but does not verify that moves are legal given the board state implied by the seed sequence. In preliminary checks we found that even a normally prompted model, rarely produced fully legal continuations, which we attribute to model scale. Illegal but syntactically valid moves may therefore pass into the training data, introducing task-level noise that could partially explain the weaker transmission observed in chess compared to number generation.

Finally, our extensions are limited in breadth. They cover two new preference categories, one new task, and one new model family. Our claims about where subliminal learning does and does not generalize should be read as suggestive rather than exhaustive. A wider set of traits, tasks, and models would be needed to map the boundary more precisely.


\bibliography{bib/custom}

@article{cloud2026language,
  title={Language models transmit behavioural traits through hidden signals in data},
  author={Cloud, Alex and Le, Minh and Chua, James and Betley, Jan and Sztyber-Betley, Anna and Mindermann, S{\"o}ren and Hilton, Jacob and Marks, Samuel and Evans, Owain},
  journal={Nature},
  volume={652},
  number={8110},
  pages={615--621},
  year={2026},
  publisher={Nature Publishing Group UK London}
}

@inproceedings{hinton2014distilling,
  title={Distilling the Knowledge in a Neural Network},
  author={Hinton, G},
  booktitle={Deep Learning and Representation Learning Workshop in Conjunction with NIPS},
  year={2014}
}

\newpage
\appendix

\section{Experimental Setup Details}
\label{sec:appendix}
We reuse the distillation code base of \citet{cloud2026language} with minimal adaptation\footnote{Available at \url{https://github.com/MinhxLe/subliminal-learning}}. This appendix describes the specific settings of our reproduction and extensions.

\subsection{Distillation pipeline}
For each trait category (animal, actor, politician) and base model (Qwen2.5-7B, Gemma3-4B, Ministral8B), we: 
\begin{enumerate}
    \item create a teacher by conditioning the reference model on a system prompt naming the target entity (e.g., ``You love owls. You think about owls all the time. Owls are your favorite animal. Imbue your answers with your love for the animal.'')
    \item sample completions from the teacher on prompts unrelated to the trait (number sequences or chess continuations)
    \item apply a filter rule, to remove completions that reference the trait explicitly or subtly
    \item fine-tune a student, initialized from the same reference model, on the filtered completions via supervised fine-tuning
\end{enumerate}

A matched \emph{regular-FT} control student is trained identically on completions sampled from an unprompted reference model of the same family.

\subsection{Dataset generation}
For both tasks, teachers continue a seed sequence of items and are instructed to generate up to 10 further items, sampled at temperature 1.0. Each teacher produces 30,000 completions, which are filtered and downsampled to 10,000 completions to hold dataset size constant across teachers. The 10,000 fine-tuning examples are shared across all fine-tuning seeds of a given trait; only the fine-tuning seed itself varies, drawn from $\{1,2,3,4,5\}$.

\subsubsection{Numbers task}
Number sequences are generated following the prompt templates of \citet{cloud2026language}. Filtering in this case focuses on formatting (numeric-only, correctly delimited, matching the requested count). For the digit-length ablation (\S4.2), the same procedure is repeated with the answer space restricted to $[0,9]$ (1-digit) or $[10,99]$ (2-digit).

\subsubsection{Chess task}
For chess move generation, the seed sequence consists of real opening half-moves (SAN notation) sampled from the public \texttt{lichess-org/chess-openings} database (approximately 3,800 distinct opening lines). Completions are filtered for valid SAN formatting and rejected if malformed or if they exceed the requested move count. As noted in the Limitations, board-state legality of the continuation is not verified.

\subsection{Fine-tuning configuration}
Students are fine-tuned with LoRA (rank $r=8$, $\alpha=8$, applied to all attention and MLP projection matrices: \texttt{q\_proj}, \texttt{k\_proj}, \texttt{v\_proj}, \texttt{o\_proj}, \texttt{gate\_proj}, \texttt{up\_proj}, \texttt{down\_proj}) via Unsloth, using the Hugging Face TRL \texttt{SFTTrainer} with completion-only loss masking (the prompt is not trained on). Training uses the AdamW optimizer with learning rate $2\times10^{-4}$, a linear schedule with 5 warmup steps, per-device batch size 22 with 3 gradient-accumulation steps (effective batch size 66), gradient clipping at norm 1.0, and 3 epochs. These hyper-parameters are identical across Qwen2.5-7B, Gemma3-4B, and Ministral8B. Training uses bf16 precision where supported by the GPU and fp16 otherwise, with gradient checkpointing. 

\subsection{Evaluation protocol}
Students are evaluated with 50 neutral prompt variants per trait category (animal, actor, politician) that never mention the trait (e.g., ``In one word, what is your favorite animal?''), each sampled 100 times at temperature 1.0 (5,000 completions per condition per model). We report the rate at which the target entity appears in the response. 

Completion health is assessed with Llama3.1-8B as an independent judge model, chosen from a different model family than any evaluated student to reduce self-preference bias.

\subsection{Statistical analysis}
Unless otherwise stated, error bars and confidence intervals are computed across 5 fine-tuning seeds per condition. The seed sets the student's LoRA adapter initialization and the training-data shuffling order (both controlled by a single seed value passed to Unsloth/TRL); it does not resample the teacher-generated dataset, which is fixed per trait (dataset generation uses a fixed seed of 42) and shared across all 5 fine-tuning seeds. Confidence intervals are 95\% intervals based on a $t$-distribution over seeds. 

Following \citet{cloud2026language}, we do not correct for multiple comparisons across entities: a narrow, non-overlapping confidence interval for a single entity should not be read as an independently significant test for that entity. Our claims instead concern whether transmission occurs broadly across entities, traits, tasks, and models, with each entity serving as one replication rather than an independent hypothesis test.

\subsection{Compute resources}
Across the full grid (3 base models $\times$ 2 primary tasks $\times$ 3 trait categories $\times$ 5 seeds, plus the Qwen-only digit-length ablation), we ran approximately $1,094$ (model, task, trait, seed) pipeline cells, each comprising dataset generation, LoRA fine-tuning, and evaluation. This amount to approximately 518 GPU-hours on NVIDIA RTX 4090s (24GB), executed across multiple GPUs in parallel (up to 7 concurrently).

\end{document}